\documentclass[hidelinks, conference]{IEEEtran}
\IEEEoverridecommandlockouts
\PassOptionsToPackage{bookmarks={false}}{hyperref} 
\usepackage[a-1b]{pdfx}

\usepackage{cite}
\usepackage{stfloats}
\usepackage{amsmath,amssymb,amsfonts}
\usepackage{adjustbox}
\usepackage{algorithmic}
\usepackage{graphicx}
\usepackage{textcomp}
\usepackage{xcolor}
\usepackage{soul}
\usepackage{caption}

\usepackage{subfig}
\def\BibTeX{{\rm B\kern-.05em{\sc i\kern-.025em b}\kern-.08em
    T\kern-.1667em\lower.7ex\hbox{E}\kern-.125emX}}
    
\begin{document}

\newcommand{\SubItem}[1]{
    {\setlength\itemindent{15pt} \item[-] #1}
}
\newcommand{\bh}[1]{\textcolor{red}{Comment: #1}}



\title{\huge {ADL-ID: Adversarial Disentanglement Learning Network for RF Fingerprinting Temporal Domain Adaptation}}

\title{\huge {ADL-ID: Adversarial Disentanglement Learning for Wireless Device Fingerprinting Temporal Domain Adaptation}}


\author{\IEEEauthorblockN{Abdurrahman Elmaghbub, Bechir Hamdaoui and Weng-Keen Wong}
\IEEEauthorblockA{\textit{School of Electrical Engineering and Computer Science},
\textit{Oregon State University}\\
\{elmaghba, hamdaoui, wongwe\}@oregonstate.edu}

}

\maketitle
\begin{abstract}
As the journey of 5G standardization is coming to an end, academia and industry have already begun to consider the sixth-generation (6G) wireless networks, with an aim to meet the service demands for the next decade. Deep learning-based RF fingerprinting (DL-RFFP) has recently been recognized as a potential solution for enabling key wireless network applications and services, such as spectrum policy enforcement and network access control. The state-of-the-art DL-RFFP frameworks suffer from a significant performance drop when tested with data drawn from a  domain that is different from that used for training data. 
In this paper, we propose ADL-ID, an unsupervised domain adaption framework that is based on adversarial disentanglement representation to address the temporal domain adaptation for the RFFP task. Our framework has been evaluated on real LoRa and WiFi datasets and showed about $24\%$ improvement in accuracy when compared to the baseline CNN network on short-term temporal adaptation. It also improves the classification accuracy by up to $9\%$ on long-term temporal adaptation. Furthermore, we release a $5$-day, $2.1$TB, large-scale WiFi 802.11b dataset collected from $50$ Pycom devices to support
the research community efforts in developing and validating robust RFFP methods.


\end{abstract}


\section{Introduction}

An important highlight of evolving next-generation wireless networks is security and privacy, which have been somehow overlooked, to some extent, in previous generations~\cite{dang2020should}. While traditional security relies heavily on the upper-layer mechanisms, it is essential to complement these mechanisms with unclonable, physical-layer security technologies to increase the security robustness of such systems. As an  efficient physical-layer security mechanism, RF fingerprinting (RFFP) technology allows for device identification based on unique fingerprints inherited in the RF signal. Due to the imperfection of the manufacturing process, circuit components in the signal path introduce a variety of random hardware impairments such as I/Q mismatch, phase noise, and non-linearity \cite{elmaghbub2020widescan} that collectively impact the transmitted signal, constituting a unique RF Fingerprint (RFF). Deep Learning (DL) has reduced the need for preprocessing and expert-defined feature extraction techniques, building state-of-the-art RFFP frameworks~\cite{merchant2018deep,hamdaoui2020deep}. However, most of these rely on the assumption that the training and testing sets are drawn from the same distribution, which falls short of the conditions of realistic RF scenarios. 

We define domain portability as the ability of a model to maintain its training domain performance when applied to new domains. In the RFFP context, the domain refers to the settings in which data has been collected. This includes time, channel, receiver, protocol, and environment. Hence, any considerable change in the training settings yields a different domain. In \cite{elmaghbub2021lora,hamdaoui2022deep}, we conducted an experimental study on LoRa devices, which disclosed the sensitivity of DL-RFFP to domain changes. In that work, the models (i) fail to maintain their high accuracy when channel conditions change and (ii) completely lose their classification ability when the protocol configuration or receiver changes. This lack of portability represents a substantial impediment to the implementation of DL-based RFFP methods in real-world applications. 

The complexity of the RFFP problem comes from the fact that we are dealing with two layers of obscurity: the fingerprint interactions in both circuit and channel levels and the feature vector learned by the deep learning network. The lack of behavioral comprehension of RFFs and deep learning networks opens up the floor to several hypotheses to explain the exposed sensitivity of deep learning models when the domain changes. It could be that the fingerprints themselves change as the domain changes; hence, identifying devices based on their old/learned fingerprints yields poor accuracy in cross-domain validation. This assumption disqualifies these
features as fingerprints since authentication schemes require them to be robust and permanent. More reasonably, the feature vector learned by the network includes both RFFs and a domain-specific component which leads to confusion when the latter component changes as the domain changes. Most of the proposed solutions adapted the latter assumption in their endeavors. Some works have focused on eliminating the background effects from the raw signal to improve its domain generalization. One such example is the use of channel equalization to improve the performance by eliminating channel
influences \cite{2019deepradioid}, but this is a partial solution since they
eliminate many beneficial spurious emissions in the passband, which reduces the user capacity of the RFFs \cite{rajendran2022rf}. On the other hand, DeepLoRa \cite{al2021deeplora}, a data augmentation technique based on the ITU-R channel models, was proposed to counter the degradation introduced by the wireless channel. The best performance reported by DeepLoRa was $36\%$ with only $5$ devices being classified, achieved when the model is trained on one day and tested on another day. This accuracy drops to $8\%$ when the number of devices goes to $50$.  

More recently, the portability problem of the RFFP has been formulated as a domain adaptation problem \cite{redko2020survey,gaskin2022tweak}, allowing us to leverage the advancements of this branch of transfer learning in the RF context. This formulation assumes that the source and target domains have slightly different distributions.
In~\cite{wang2022specific}, a multi-discrepancy domain adaptation method has been proposed to minimize the intraclass discrepancy across the source and target domains while maximizing the inter-class discrepancy, leading to $15\%$ improvement over the ResNet network. Adversarial domain adaptation \cite{wang2021specific} has also been proposed to reduce the influence of domain features on the RFFs by maximizing the domain discrimination loss, resulting in an enhancement in the portability of the network. However, the target domains on the two mentioned domain adaptation methods have been created via simulation by introducing white Gaussian noise to the source domain dataset to create several target domains with different signal-to-noise ratio conditions. These simulation settings are far from realistic RF scenarios where multiple settings change at the same time. Addressing multiple varying settings, Tweak~\cite{gaskin2022tweak} combined the metric learning with a lightweight calibration method to address the portability problem in hardware, channel, and configuration dimensions, separately. Tweak shows a considerable performance improvement in each of these dimensions, but needs to be recalibrated each time the target domain changes.

Another exciting domain adaptation category is disentangled representation learning, an unsupervised learning technique that separates (disentangles) features into disjoint parts of the underlying representation\cite {higgins2018towards}. Motivated by the intuition that the RFFs are entangled with domain-specific features and inspired by the disentanglement concept~\cite{bousmalis2016domain}, we propose ADL-ID,  a domain adaptation framework that integrates disentanglement representation learning with adversarial learning to address the temporal portability problem in RFFP. Our framework factors the feature vector into a device-specific component (RFFs) and a domain-specific component. Then, it uses only the device-specific component, which contains domain-invariant features, as input to the classifier. Our proposed framework, tested using real WiFi dataset (detailed in Section~\ref{sec:testbed}) and LoRa datasets (described in~\cite{elmaghbub2021comprehensive}), provides a high-performance gain, $20\%-24\%$, in short-term-varied domain adaptation. We are also excited to release our large-scale WiFi 802.11b dataset of $50$ Pycom devices to the research community to support the efforts toward enabling RFFP portability. This is the first public WiFi 802.11b RFFP dataset, which is common among low-rate IoT devices. Our testbed includes the same devices used before to build and release the LoRa dataset~\cite{elmaghbub2021comprehensive}, which extends the opportunity to analyze other domain dimensions, such as protocol configurations and hardware receivers. Our main contributions are as follows:
\begin{itemize}
    \item Proposed an adversarial-based disentanglement learning network to address the RFFP temporal portability issue. 
    \item Released the first large-scale WiFi 802.11b RFFP dataset of $50$ Pycom devices over $5$ consecutive days for both indoor and outdoor scenarios. 
    \item Provided novel insights into the performance of RFFP methods due to short-term and long-term temporal domain variations using both WiFi and LoRa datasets. 
\end{itemize}

The rest of the paper is organized as follows. Section \ref{sec:testbed} describes the device testbed and WiFi dataset. Following that, Section~\ref{temporal} reveals the investigation on short-term variations performance. Then, our framework is explained in Section~\ref{proposed} while our results are presented and discussed in Section \ref{evaluation}. The paper is concluded in Section \ref{Conclusion}.

\label{intro}

\section{Testbed and WiFi Dataset Description}
\label{sec:testbed}
\begin{figure}[t]
    {%
       \includegraphics[width=0.48\textwidth, height = 0.30\textwidth]{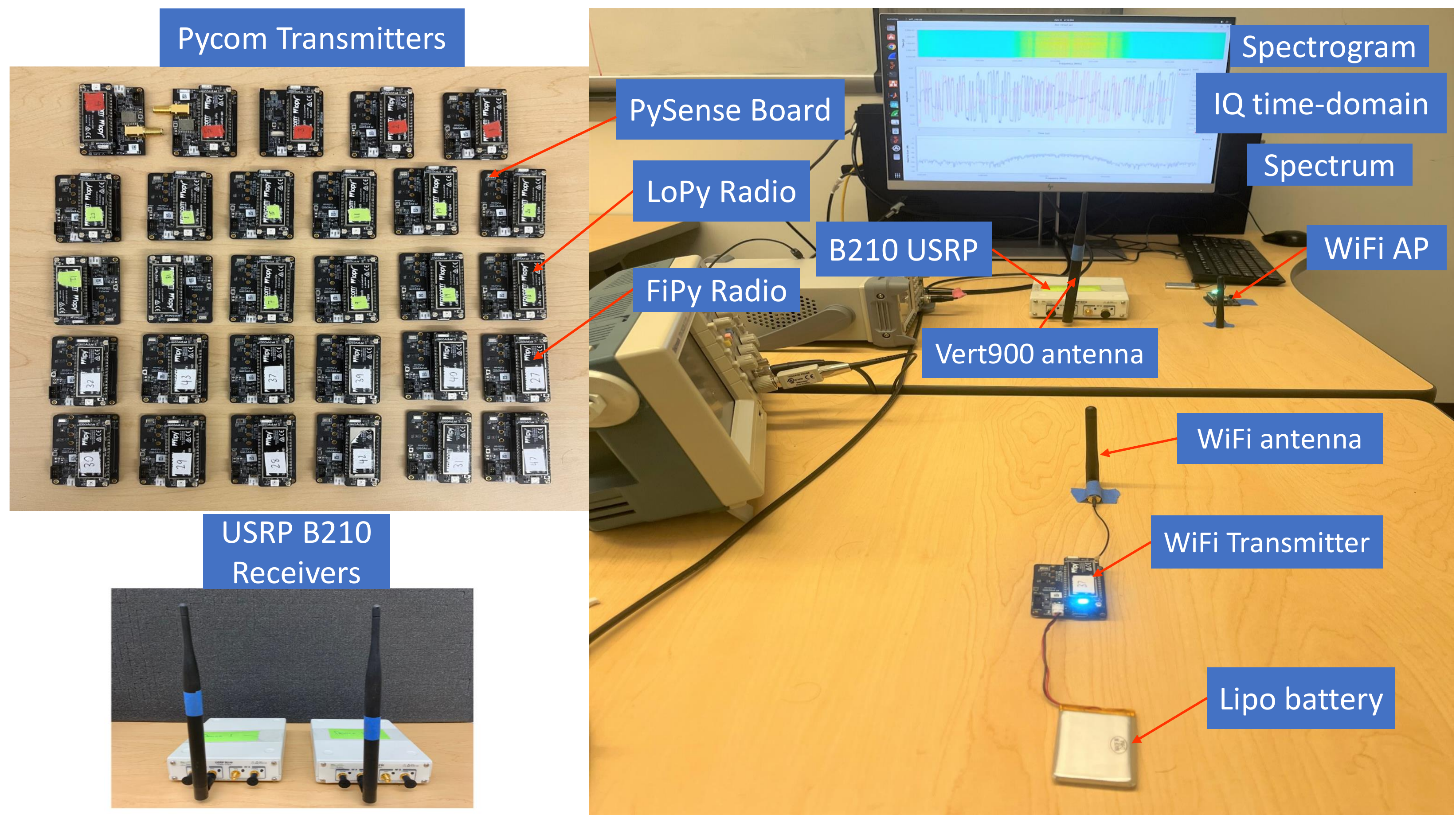}
     }
     \caption{50-device Pycom WiFi testbed}
     \label{fig:testbed}
\end{figure}

In this section, we begin by describing the hardware, software, and protocol components of the tested used for collecting our dataset. Then, we describe the collected WiFi dataset. The description and download information of the LoRa dataset that we also used in our evaluation can be found in \cite{elmaghbub2021comprehensive,elmaghbub2021lora}.

\subsection{Testbed Description}
\subsubsection{Hardware}
Our testbed, shown in Fig.~\ref{fig:testbed}, consists of $50$ Pycom devices ($25$ FiPy \& $25$ LoPy boards on top of $50$ PySense extension boards). These boards mainly include ESP$32$-D$0$WDQ, Semtech SX$1276$, and Sequans Monarch chips to support WiFi, Bluetooth, LoRa, Sigfox, and LTE networks, respectively. On the reception side, we used an Ettus USRP (Universal Software Radio Peripheral) B$210$ with a VERT$900$ antenna for the data acquisition. The RF-front of the USRP was configured with a center frequency of $2.412$GHz, a $20$MHz Bandwidth, and a $25$MSps sample rate. We used lipo batteries to power the Pycom devices. Each Pycom device was connected to a dedicated WiFi antenna and configured to transmit IEEE802.11b packets in the $2$GHz spectrum.

\subsubsection{Software}
Pycom devices are based on FREERTOS operating system. To perform transmission, we programmed and configured our Pycom boards using MicroPython, an efficient implementation of Python3 composed of a subset of standard Python libraries and optimized to run on microcontrollers and constrained environments. Also, we used Pymakr plugin as a REPL console that connects to Pycom boards to run codes or upload files. 
To perform reception, we used the GNURadio software, a real-time signal processing graphical tool, along with USRP Hardware Driver (UHD) to set up and configure the USRP receiver to capture WiFi transmissions, plot their time and frequency domains, 
and store the collected samples in their files.

\subsubsection{Protocol}
We used the WiFi 802.11b protocol with the high-rate direct-sequence spread-spectrum (HR/DSSS) physical layer. Transmitters were configured to transmit at $1$Mbps, with a carrier frequency of $2.412$GHz and $20$MHz bandwidth. 

\subsection{Dataset Description}
Our WiFi dataset contains $2.1$TB of WiFi transmissions of $50$ Pycom devices captured over $5$ consecutive days in both indoor (laboratory) and outdoor environments. For each day, we capture $5$ transmissions from each device in a round-robin fashion, where each capture consists of $50$M complex-valued I/Q samples. The time-gap between two consecutive captures of the same device is $5$mins. Transmitters were located about $5$ meters away from the access point and the receiver, with a clear line of sight. The collected signals are down-converted by the USRP to the baseband and then stored in the form of I/Q data in both time and frequency domain representations. For each capture, associated metadata compliant with the
latest SigMF specifications and additional field extensions for usability are recorded.

More details and use cases of the WiFi dataset can be found in~\cite{elmaghbub2021lora}. The datasets can be downloaded from NetSTAR Lab at 
\href{https://research.engr.oregonstate.edu/hamdaoui/datasets/}{\color{blue}{http://research.engr.oregonstate.edu/hamdaoui/datasets}}.

\section{The Temporal Portability Problem}
\label{temporal}
\begin{figure}
     \subfloat[LoRa Indoor Scenario.\label{subfig-1:phase1}]{%
       \includegraphics[width=0.48\textwidth, height = 0.20\textwidth]{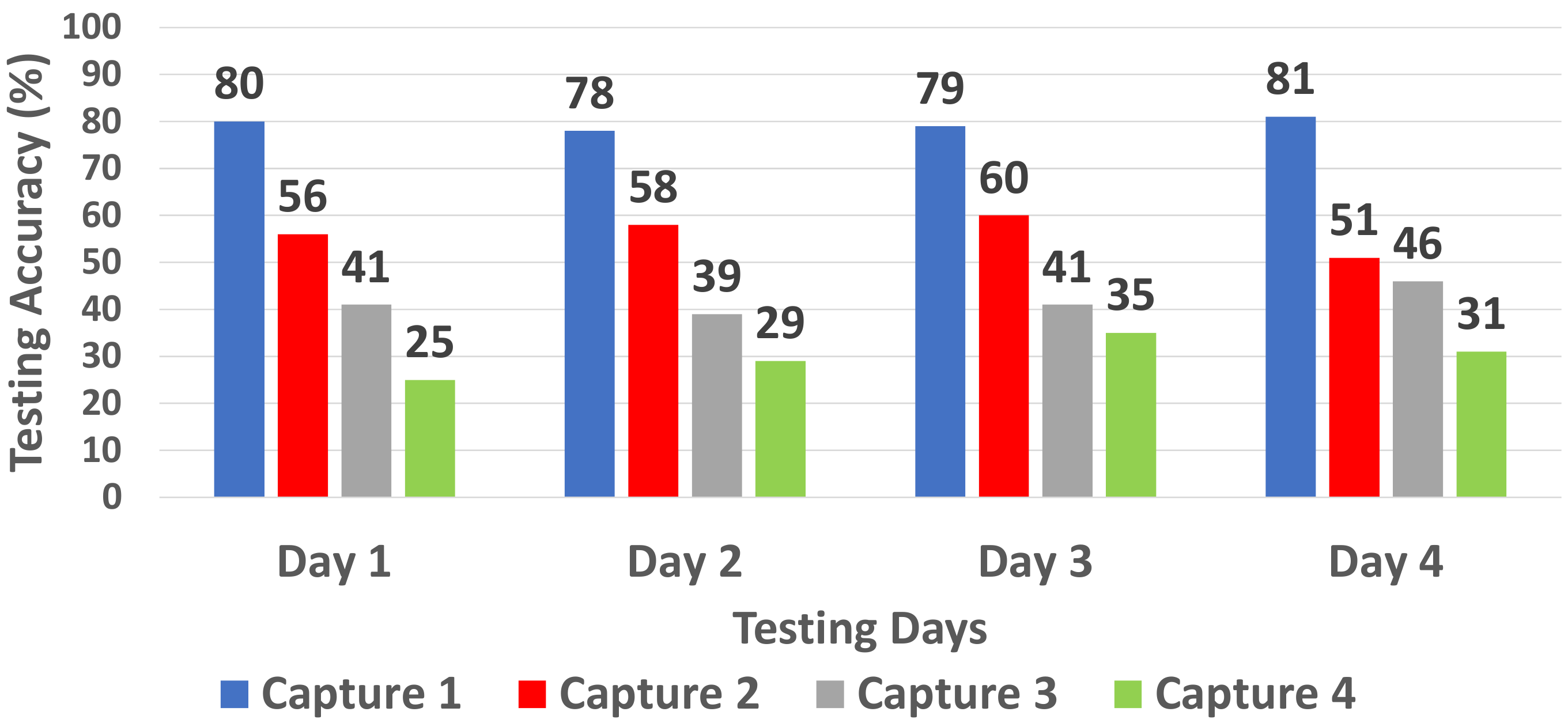}
     }
    \hspace{0.00001cm}
     \subfloat[LoRa Outdoor Scenario.\label{subfig-2:phase2}]{%
       \includegraphics[width=0.48\textwidth, height = 0.20\textwidth]{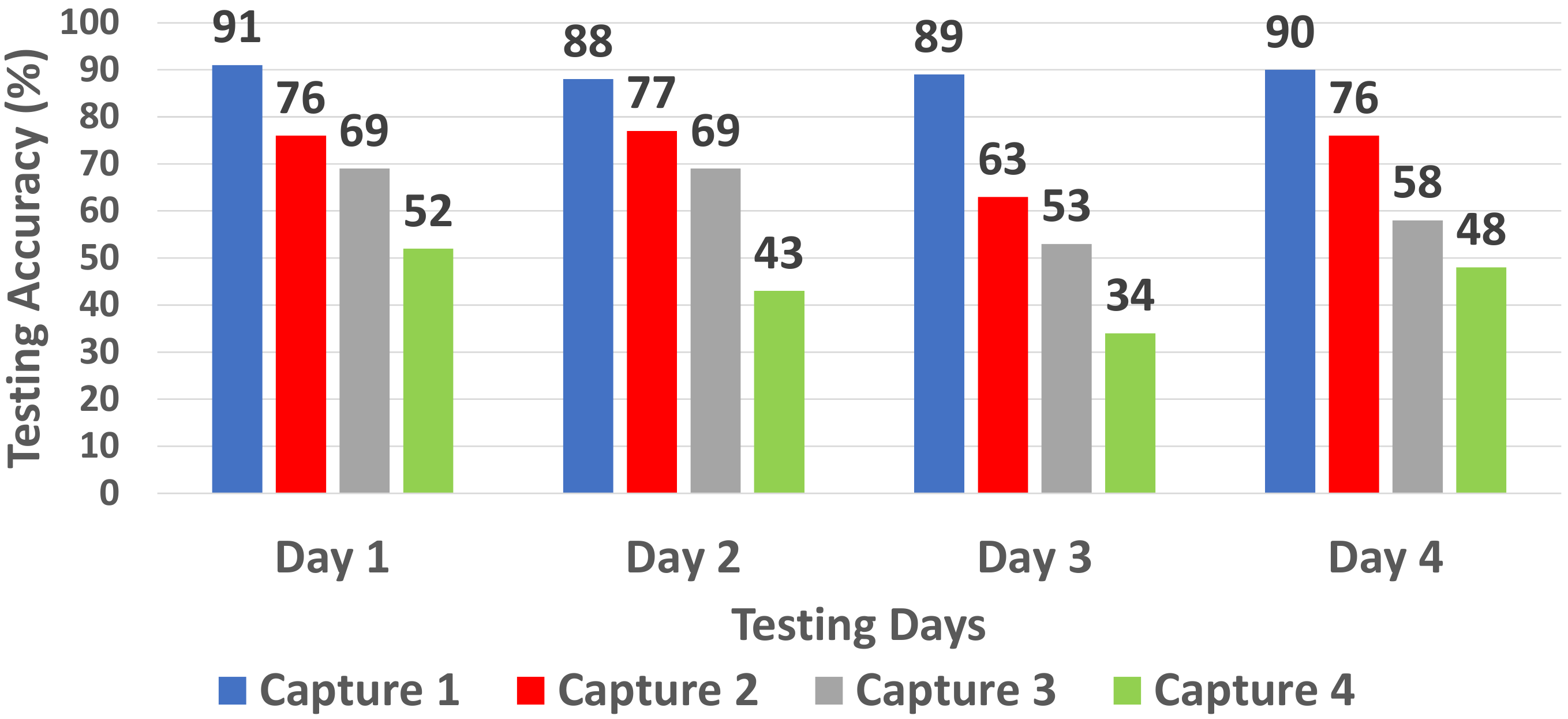}
     }
     \caption{Performance in short-term variations of LoRa captures where the time gap between two consecutive captures is $5$mins. The CNN model was trained on captures $1$ of each day and tested on captures $1$, $2$, $3$, and $4$.}
     \label{fig:lora_cap}
\end{figure}

Several works \cite{elmaghbub2021lora,al2020exposing, al2021deeplora} consider the days as the change unit in the temporal dimension of the domain portability problem. Hence, they analyze the performance of the models when the training and testing data are from different days. However, there is nothing special about the 24 hours as far as the RF fingerprints are concerned. We postulate that the data distribution of fingerprints changes dramatically even within the same day. To validate this hypothesis, we used our LoRa and WiFi datasets in which we have several captures of the same device within the same day. Starting with the LoRa dataset \cite{elmaghbub2021comprehensive}, the time gap between two consecutive captures is $5$ minutes with $5$m distance between transmitters and the receiver. The captures were collected in indoor and outdoor scenarios, and the devices have not been rebooted between the captures. We trained a 6-Layer CNN network \cite{elmaghbub2021lora} using capture $1$ and tested it on data from captures $1$, $2$, $3$, and $4$ of the $4$ days. Surprisingly, the model's performance suffers from a huge degradation even when the time gap between the training and testing sets is as short as $5$ minutes. Furthermore, as we can see in Fig~\ref{fig:lora_cap}---depicting accuracy results when using the LoRa data, the performance continues to degrade as we move further in time from the training domain. This trend manifests itself in the four days and both scenarios. 

\begin{figure}
     \subfloat[WiFi Indoor Scenario.\label{subfig-1:phase1}]{%
       \includegraphics[width=0.48\textwidth, height = 0.20\textwidth]{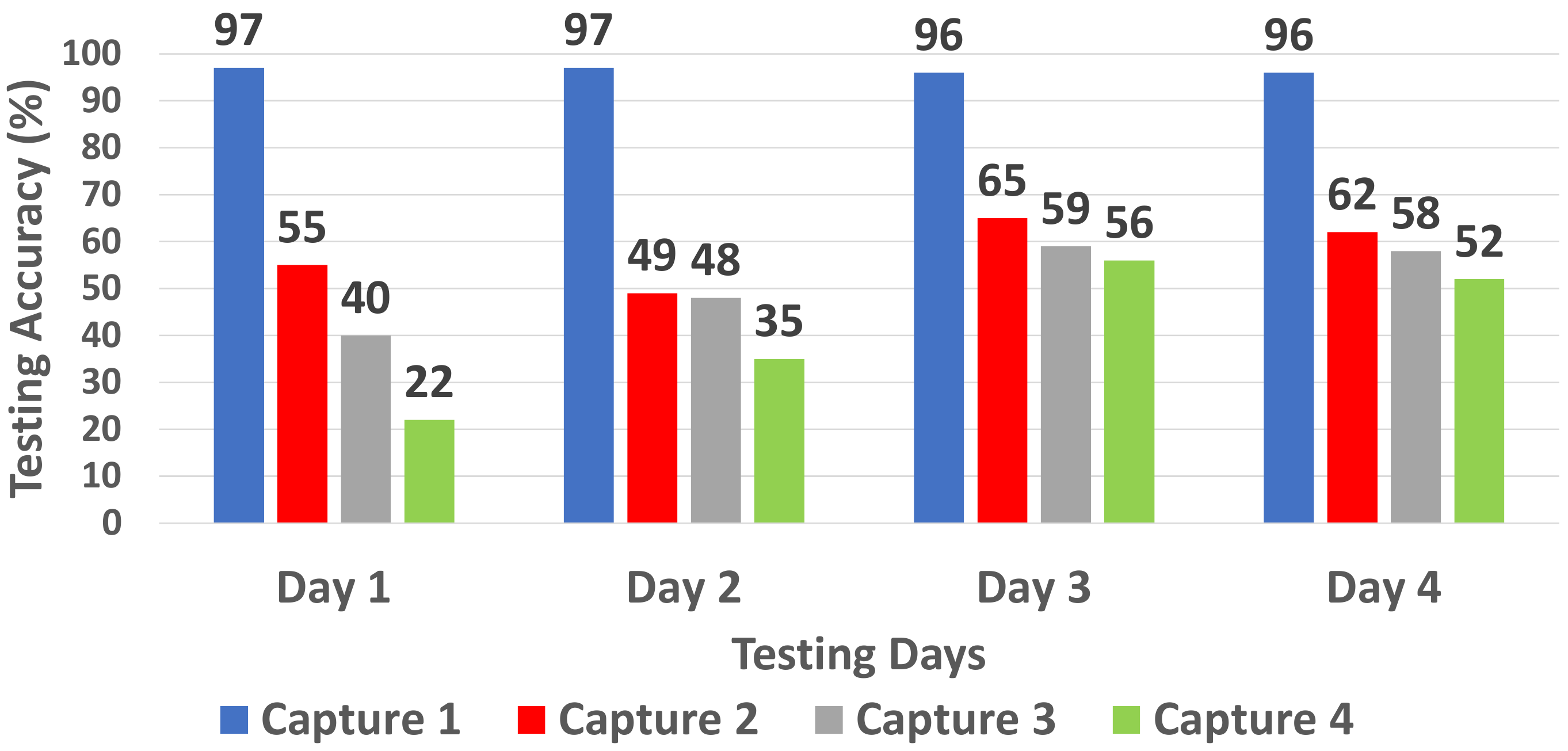}
     }
    \hspace{0.00001cm}
     \subfloat[WiFi Outdoor Scenario.\label{subfig-2:phase2}]{%
       \includegraphics[width=0.48\textwidth, height = 0.20\textwidth]{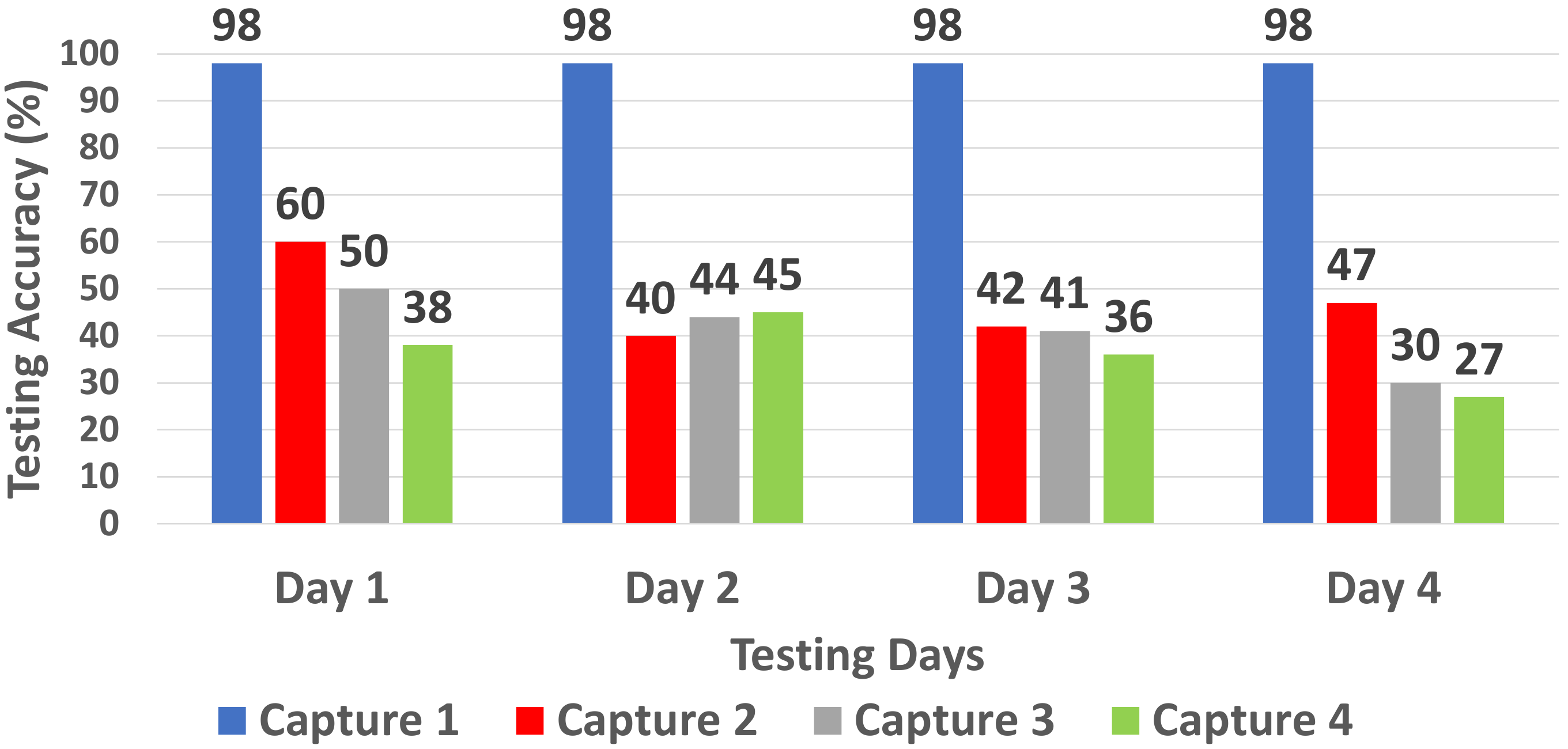}
     }
     \caption{Performance in short-term variations of WiFi captures where the time gap between two consecutive captures is $5$mins. The CNN model was trained on captures $1$ of each day and tested on captures $1$, $2$, $3$, and $4$.}
     \label{fig:wifi_cap}
\end{figure}

\begin{figure}
    {%
       \includegraphics[width=0.48\textwidth, height = 0.20\textwidth]{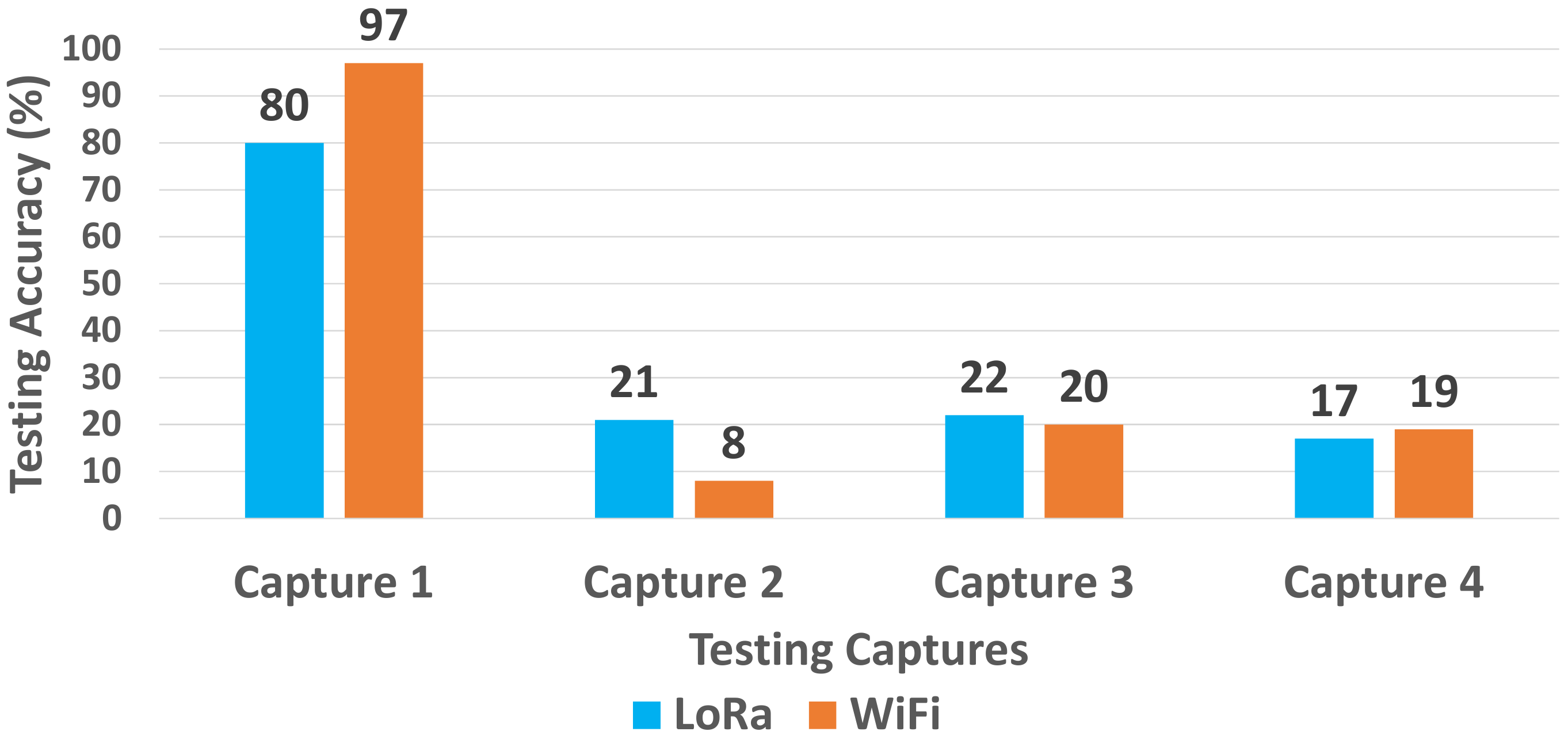}
     }
     \caption{Performance in long-term variations on LoRa (1st column) and WiFi (2nd column) captures where the captures are collected on $4$ different days. The CNN model was trained on capture $1$ and tested on captures $1$, $2$, $3$, and $4$. }
     \label{fig:long-term}
\end{figure}

To confirm our observation, we carried out the same investigation in our 50-device WiFi dataset, which also includes $5$ captures of the same devices within the same day for $5$ consecutive days, where the time gap is similar to that in the LoRa dataset. The difference is that the devices here have been rebooted between the captures. Results in Fig.~\ref{fig:wifi_cap} show that the WiFi dataset is no different, and the model also suffers from short-term variations on the WiFi dataset. These results consolidate our hypothesis that even a slight departure from the network's training domain can introduce confusion to the classifier and hurt its performance severely. 

We also notice that on most of the days, the most significant drop rate in performance occurs in the transition from capture $1$ to capture $2$, and that rate decreases as we move away from the training domain. We claim that the drop rate decreases until it reaches a point when the time gap does not matter anymore; i.e., aging has no to little effect. Hence, with respect to domain adaptation performance, $2$-day or $10$-day time separation between the source and target domains has a similar effect. Some of that can be seen in Fig.~\ref{fig:long-term}, which indicates that the effect on the different days is similar.
We refer to this effect as the effect of long-term variations. As expected, Fig.~\ref{fig:long-term} shows that the drop in the network's performance due to the long-term variations is more severe than the previous scenario as the accuracy drops from around $90\%$ to $20\%$, when the model is trained on capture $1$ and tested on captures $2$, $3$, and $4$, which have been collected on day $2$, $3$, and $4$, respectively.

\section{The Proposed Framework: ADL-ID}
\label{proposed}
\begin{figure*}
    \centering
    \includegraphics[width=2.05\columnwidth, height=0.77\columnwidth]{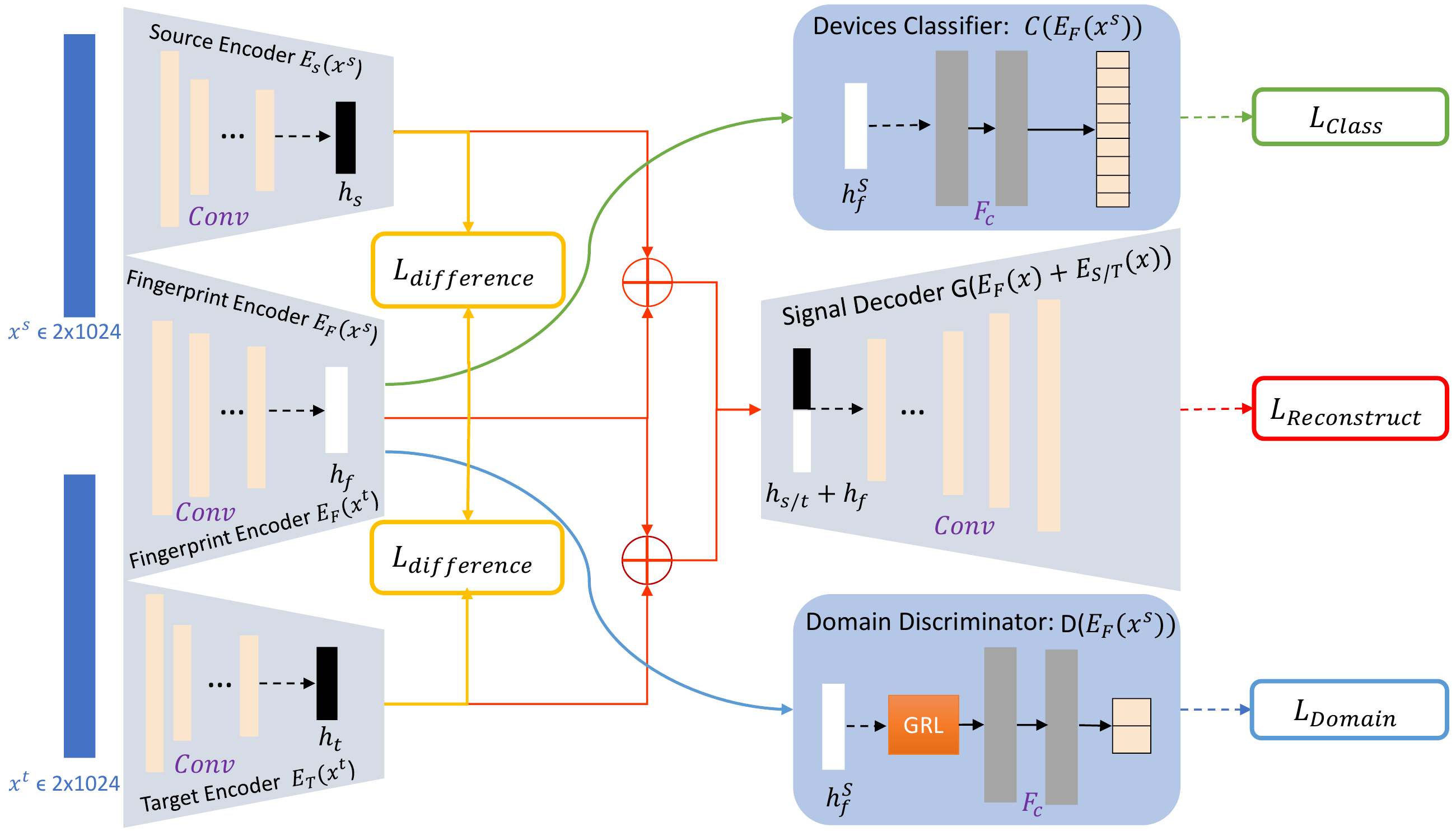}
    \caption{Architecture of the proposed unsupervised adversarial-based disentanglement learning network for the RFFP task.}
    \label{fig:frameworf}
\end{figure*}
The proposed framework, shown in Fig.~\ref{fig:frameworf}, aims to train an encoder, called fingerprint encoder, $E_F$, to extract only the device-specific feature vector (RFFs) and feed it to the cross-domain device classification task. The intuition here is that the RFFs are shared across domains while the background and domain artifacts are different from one domain to another. This is achieved by having three encoders: fingerprinting encoder, $E_F$, source encoder, $E_{S}$, and target encoder, $E_{T}$ that are trained to model different components of the domain representation. Using labeled source domain data, $X_S = \{(x_s^i, y_s^i)\}^{N_s}_{i=0}$, and unlabeled target domain data, $X_T = \{(x_t^i)\}^{N_t}_{i=0}$, the fingerprint encoder learns to extract discriminative and shared features across the source and target domains while the source and target encoders learn the features that are specific to the source and target domains, respectively. \newline
At each training iteration, the output of the fingerprinting encoder will be fed to the domain discriminator, $D(x)$, to eliminate domain-indicative features to make the RFFs similar to each others regardless of their original domains. We minimize the domain classification loss, $L_{Domain}$, while having the gradient reversal layer (GRL) \cite{ganin2016domain} in the data path.  GRL acts as an identity transformation in the forward path while reversing the gradient's sign in the backpropagation. Hence, it maximizes the confusion of the domain discriminator as the updated weights of $E_F$ result in RFFs with less domain-indicative information. To enhance the separation between the RFF and the domain-specific representations of an input frame as the output of $E_F$ and $E_S$ or $E_T$ encoders, we apply a difference loss, $L_{difference}$ that encourages the orthogonality of the two representations and pushes them further away from each other by minimizing the squared Frobenius norm of the multiplication of two matrices $H_f$ and $H_s$ or $H_t$ whose rows are $h_f$ and $h_s$ or $h_t$, respectively \cite{bousmalis2016domain}. To avoid trivial solutions from the previous losses, we use a signal decoder, $G(x)$, that decodes the combination of the RFF and domain-specific features to get back the original input frame. The reconstruction process is governed by the reconstruction loss, $L_{Reconstruct}$, which is a scale–invariant mean squared error (SI-MSE) loss that experimentally proves to be more effective for disentanglement representations \cite{eigen2014depth}. Finally, $h_f^s$, the RFF representation of $X_s$ is passed to the main-task classifier with a cross-entropy loss, $L_{Class}$, that encourages the RFF representation to not just be domain-invariant and separable from the domain-specific representation, but also to be discriminative with regard to the device classification task. The total loss of our network can thus be written as
$$L_{total} = L_{Class} + \alpha L_{Reconstruct} + \beta L_{Difference} + \gamma L_{Domain}$$
where the hyperparameters, $\alpha$, $\beta$, and $\gamma$, control the strength of these losses in the overall training process. 

Each encoder in our framework consists of $6$ convolution blocks followed by a fully-connected layer. Each convolution block contains 2D convolution, batch normalization, leakyReLu, and max pooling layers. The signal decoder has also the same components in an opposite order. On the other hand, the device classifier and domain discriminator has three and two fully-connected layers, respectively, followed by a Softmax layer. We start the training with a warm up phase of $10000$ steps where only the supervised task and the reconstruction losses are used before switching to the other losses. This can help in speeding up the training process as well as leading to a better generalization \cite{wulfmeier2017addressing}.

 \section{Performance Evaluation and Discussion}
\label{evaluation}
\begin{figure}[t]
     \subfloat[LoRa Dataset.\label{subfig-1:lora_adapt_short}]{%
       \includegraphics[width=0.48\textwidth, height = 0.20\textwidth]{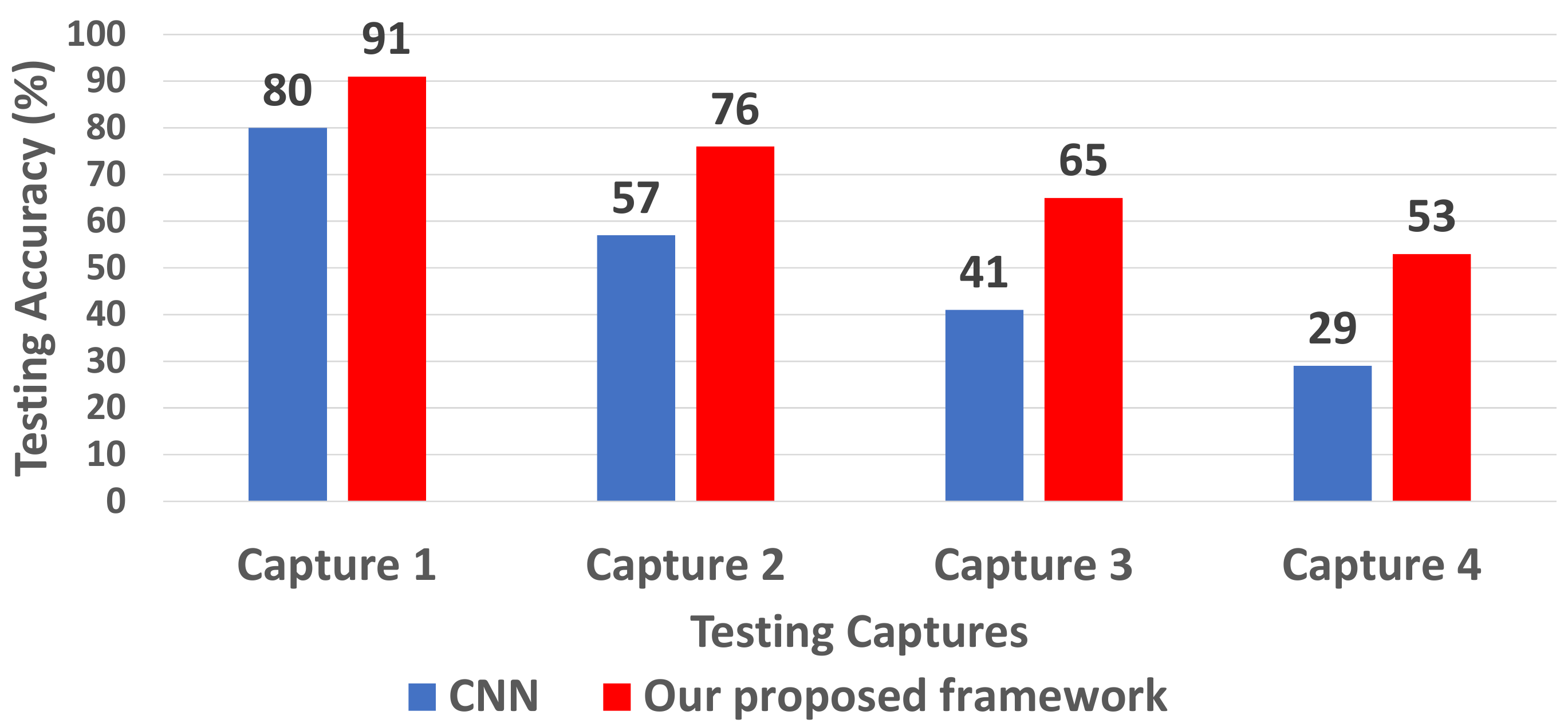}
     }
    \hspace{0.00001cm}
     \subfloat[WiFi Dataset.\label{subfig-2:wifi_adapt_short}]{%
       \includegraphics[width=0.48\textwidth, height = 0.20\textwidth]{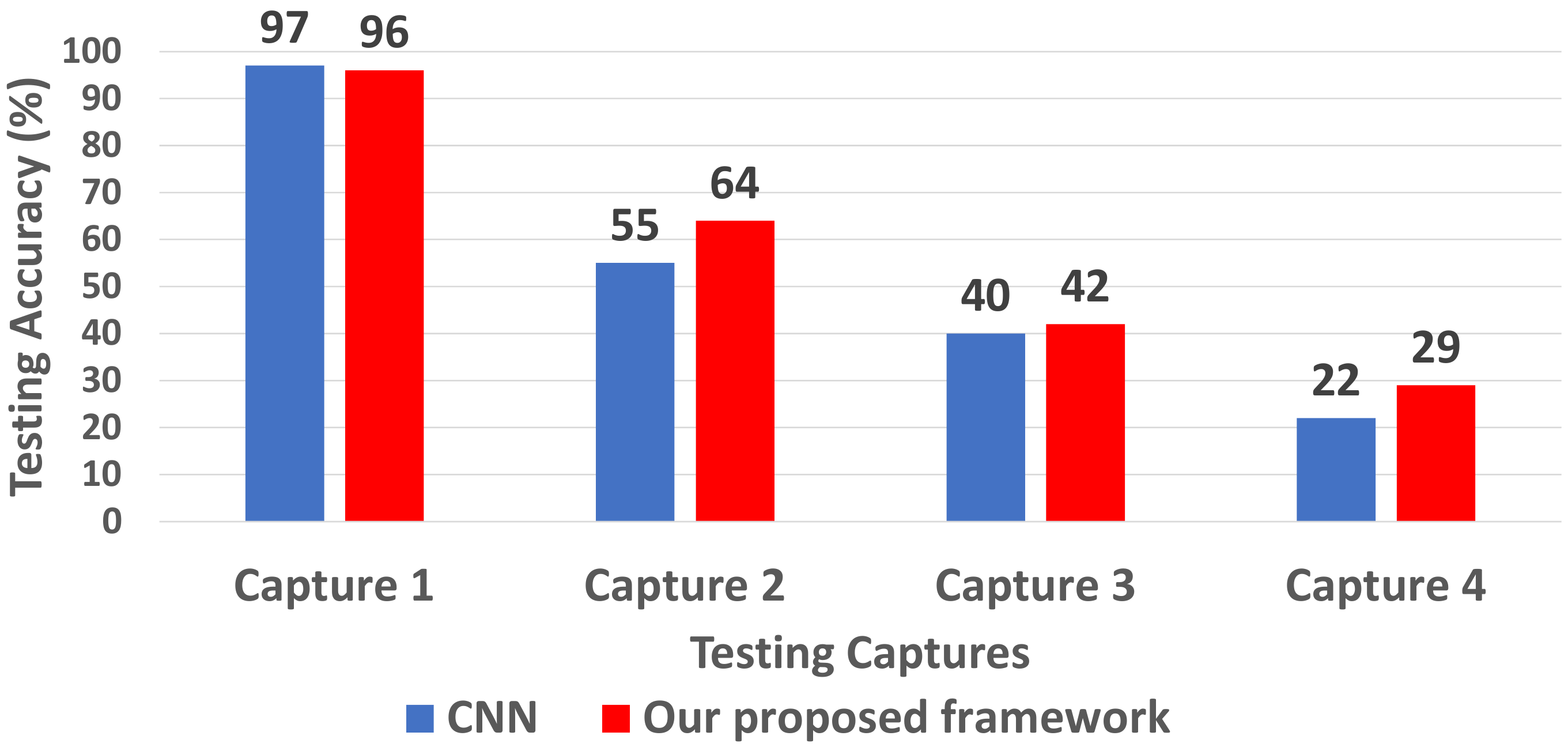}
     }
     \caption{ADL-ID vs. CNN on captures from Day $1$, Indoor scenario for both (a) LoRa and (b) WiFi. ADL-ID was trained on labeled capture $1$ data (source) and unlabeled capture $2$ data (target) and tested on captures $1$, $2$, $3$, and $4$.}
     \label{fig:short_adapt}
\end{figure}
\begin{figure}[t]
     \subfloat[LoRa Dataset.\label{subfig-1:lora_long_adapt}]{%
       \includegraphics[width=0.48\textwidth, height = 0.20\textwidth]{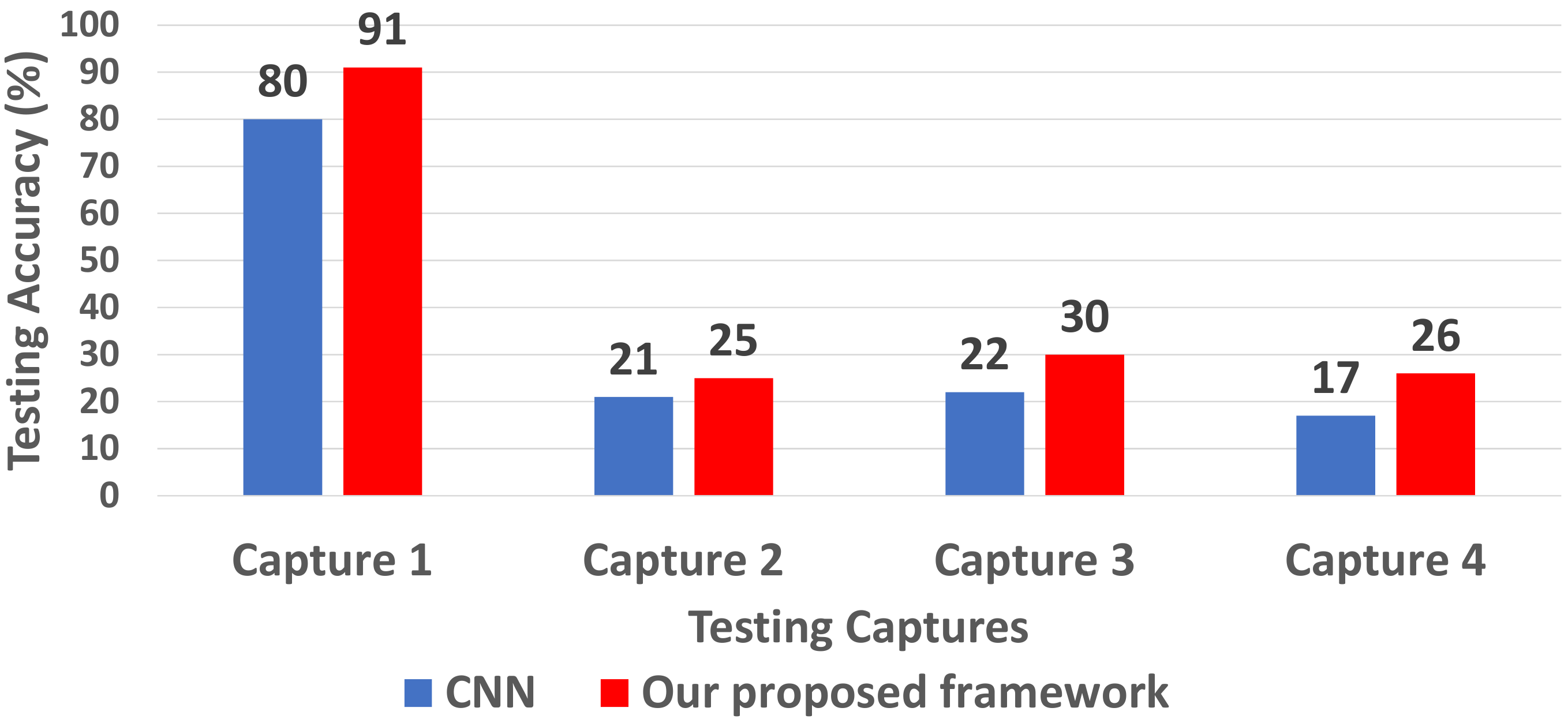}
     }
    \hspace{0.00001cm}
     \subfloat[WiFi Dataset.\label{subfig-2:wifi_long_adapt}]{%
       \includegraphics[width=0.48\textwidth, height = 0.20\textwidth]{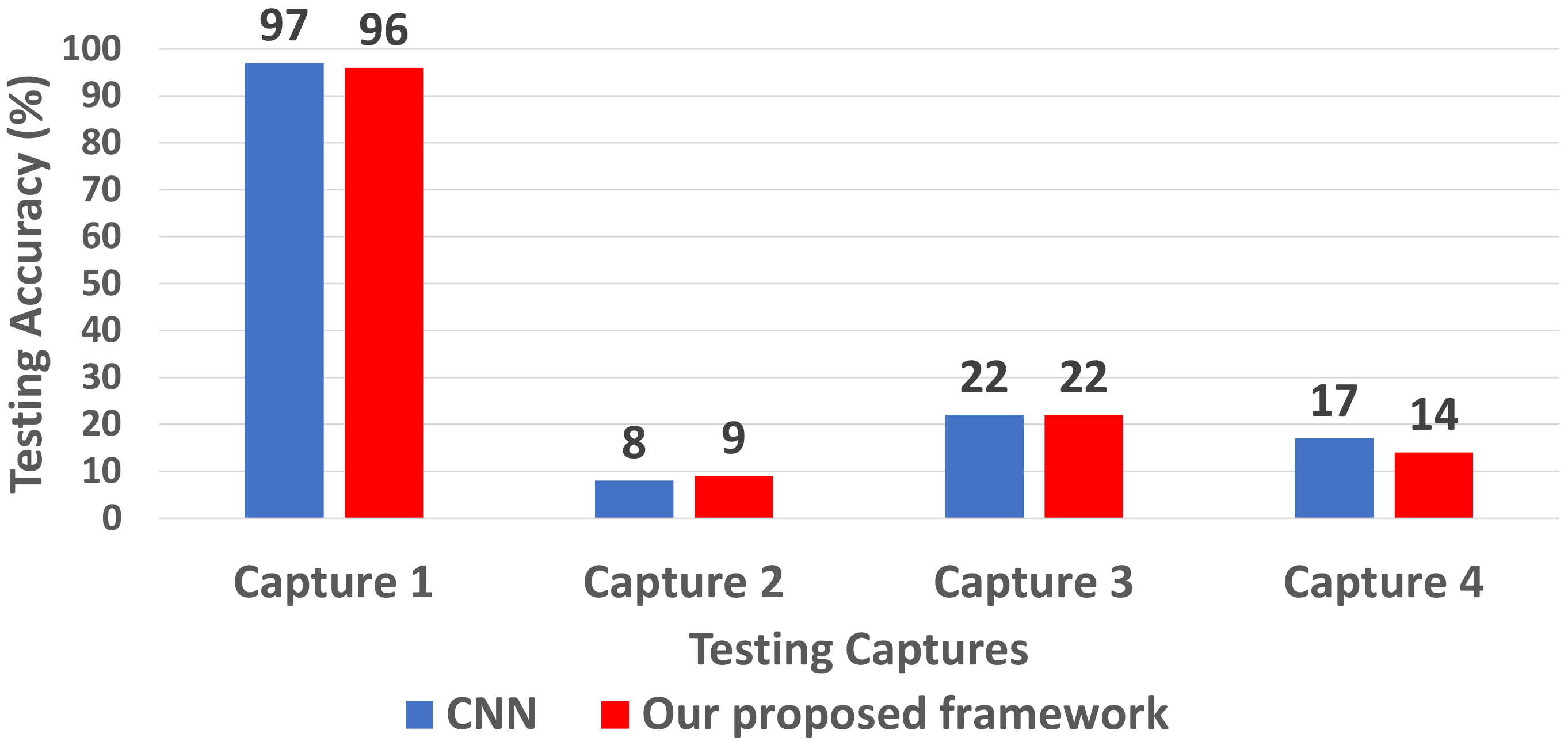}
     }
     \caption{ADL-ID vs. CNN on captures from 4 different days, Indoor scenario for both (a) LoRa and (b) WiFi. ADL-ID was trained on labeled capture $1$ data (source) and unlabeled capture $2$ data (target) and tested on captures $1$, $2$, $3$, and $4$.}
     \label{fig:long_adapt}
\end{figure}
In this section, we evaluate our proposed framework for the indoor setup of our LoRa and WiFi datasets. For each device, we used $20,000$ I/Q frames each of size $2$x$1024$, partitioned into training, validation, and testing sets. Our network has been trained for $50$ epochs on an Nvidia DGX2 node. We use two testing methodologies to run our evaluation: (i) testing on same-day captures to evaluate the short-term temporal adaptation; and (ii) testing on different-days captures to evaluate the long-term temporal adaptation.

First, we start with the short-term temporal adaption scenario. In the LoRa dataset, the time gap between the source domain and target domain is $5$mins, and the transmitters have not been rebooted between the captures. Specifically, we train our framework with labeled data from capture 1 (source domain) and unlabeled data from capture 2 (target domain) and then test it on captures $1$, $2$, $3$, and $4$, all captured on the same day with $5$mins in between. Compared to the CNN baseline, Fig.~\ref{subfig-1:lora_adapt_short} shows that ADL-ID (in red), when tested on LoRa data, provides a $20\%-24\%$ improvement in the classification accuracy, pushing the performance from $57\%$ to $76\%$ on capture $2$, from $41\%$ to $65\%$ on capture $3$, and from $29\%$ to $53\%$ on capture $4$. This improvement can be linked to the efforts of our network to minimize the presence of domain-specific features from the source and target domains in the RFF feature vector, which enhances the network's generalization. On the WiFi dataset (Fig.~\ref{subfig-2:wifi_adapt_short}), our proposed framework continues to show an improvement in accuracy with highest gain of $9\%$, increasing the accuracy from $55\%$ to $64\%$ on capture $2$, and from $40\%$ to $42\%$ on capture $3$, and from $22\%$ to $29\%$ on capture $4$. It can be noticed that the improvement percentage is higher in the LoRa dataset. We speculate that one reason could be that, in contrast to the LoRa dataset, the transmitters in the WiFi scenario were rebooted between the different captures. Consequently, a partial RF calibration has taken place, which affects the RFF distribution and stretches the distance between the source and target domains. 

Second, we evaluate our network on long-term temporal adaptation. Specifically, we consider day $1$ data as a source domain and day $2$ data as the target domain. Hence, we train the network with labeled day $1$ data and unlabeled day $2$ data and then test it on data from the $4$ days. While Fig.~\ref{fig:long_adapt} shows that our network provides an improvement over the baseline CNN network in the long-term temporal adaptation as well, it must be marked that it did not succeed in bringing the performance on long-term varying domains closer enough to the source domain's performance. A reason for this is that our framework struggles to adapt to new domains when the difference between the source and target domains is too severe, which is the case in the long-term variation scenario. This results also conveys that performance of the framework is dependent on the similarity between the source and target domains. It is worth mentioning that similar behavior has been reported when domain adaptation techniques are applied in vision applications \cite{wulfmeier2017addressing}. The variations in target domains in real-world RF applications are exceptionally more complicated compared to the vision domain. In the RFFP context, the variations in the target domain might include changes in the temporal dimension, channel characteristics, protocol configurations, and receiver hardware variations. Furthermore, these variations have made the interaction between the RFFs and domain-specific features more subtle, which results in a more difficult disentanglement process.  

\section{Conclusion}
\label{Conclusion}
We presented a large-scale WiFi 802.11b dataset, collected from 50 Pycom devices, and analyzed the performance of CNN-based RFFP on short- and long-term variations using LoRa and WiFi data. We also proposed ADL-ID, which provides up to $24\%$ improvement in short-term temporal adaptations and up to $10\%$ improvement in long-term variations over the baseline CNN. These results prove the potential of disentanglement representation techniques in enabling robust RFFP. However, more work is still needed to fully understand the RFF behaviors when the domain changes and to develop more robust RFFP methods.


\section{Acknowledgment}
\label{ack}
This research is supported in part by Intel/NSF MLWiNS Award No. 2003273. We would like to thank Intel researchers, Dr. Kathiravetpillai Sivanesan, Dr. Lily Yang, Dr. Richard Dorrance, Dr. Vesh Raj Sharma Banjade, and Dr. Hechen Wang for their constructive discussions and feedback. 

\vspace{12pt}
\color{black}
\bibliographystyle{IEEEtran}
\bibliography{IEEEexample}
\end{document}